\documentclass[conference]{IEEEtran}
\IEEEoverridecommandlockouts
\usepackage{cite}
\usepackage{float}
\usepackage{amsmath,amssymb,amsfonts}
\usepackage{algorithmic}
\usepackage{graphicx}
\usepackage{textcomp}
\def\BibTeX{{\rm B\kern-.05em{\sc i\kern-.025em b}\kern-.08em
    T\kern-.1667em\lower.7ex\hbox{E}\kern-.125emX}}
\begin{document}

\title{Deep Reinforcement Learning for Doom using Unsupervised Auxiliary Tasks}

\author{\IEEEauthorblockN{Georgios Papoudakis}
\IEEEauthorblockA{\textit{Aristotle University of Thessaloniki} \\
\textit{Electrical and Computer Engineering}\\
Thessaloniki, Greece\\
papoudak@auth.gr}
\and
\IEEEauthorblockN{Kyriakos C. Chatzidimitriou}
\IEEEauthorblockA{\textit{Aristotle University of Thessaloniki} \\
\textit{Electrical and Computer Engineering}\\
Thessaloniki, Greece \\
kyrcha@issel.ee.auth.gr}
\and
\IEEEauthorblockN{Pericles A. Mitkas}
\IEEEauthorblockA{\textit{Aristotle University of Thessaloniki} \\
\textit{Electrical and Computer Engineering}\\
Thessaloniki, Greece \\
mitkas@auth.gr}
}

\maketitle

\begin{abstract}
Recent developments in deep reinforcement learning have enabled the creation of agents for solving a large variety of games given a visual input. These methods have been proven successful for 2D games, like the Atari games, or for simple tasks, like navigating in mazes. It is still an open question, how to address more complex environments, in which the reward is sparse and the state space is huge. In this paper we propose a divide and conquer deep reinforcement learning solution  and we test our agent in the first person shooter (FPS) game of Doom. Our work is based on previous works in deep reinforcement learning and in Doom agents. We also present how our agent is able to perform better in unknown environments compared to a state of the art reinforcement learning algorithm.
%
%
\end{abstract}


\section{Introduction}
Recent advances in deep learning enabled the combination of neural networks and reinforcement learning. The first successful work in deep reinforcement learning Mnih \textit{et al.} \cite{b1}, was a Q-learning agent in which the state action value function was approximated by a deep network. In order to increase the performance and the stability of the agent, an experience replay buffer was proposed. This agent was able to achieve superhuman results in a large number of Atari games.

Despite the large success of this work, it could only be used with off-policy learning algorithms. This drawback was overcome by the introduction of an asynchronous framework Mnih \textit{et al.} \cite{b2} in reinforcement learning. This development enabled the combination of deep networks and on-policy algorithms, and resulted in the removal of the experience buffer. Additionally, the new agents were much more stable and their performance increased compared to previous works. As a result, deep reinforcement learning was able to perform decently in more complex environments, like 3D mazes.

The huge success of deep reinforcement learning resulted in an increased interest of the research community for various applications. One interesting research question is how to improve the existing algorithms in order to be able to successfully act in more complex environments like Doom, Starcraft, etc. These environments are much more complex, with a large set of states and sparse rewards. This results in a lack of a standard procedure in order to address such tasks, despite the fact that there were many successful applications of deep reinforcement learning the recent years.

This paper is focused on the creation of an agent using state of the art deep reinforcement algorithms. Additionally, inspired from previous works we propose a specific solution for the game of Doom based on three unsupervised auxiliary tasks; the value function replay, the reward prediction and the pixel control.

The structure of the paper is as follows: in section \ref{sec:theor} we discuss the theoretical background around deep reinforcement learning, while in section \ref{sec:relev_work} we refer to major works in agents for the game of Doom. In section \ref{sec:our_work} we introduce our agent and in section \ref{sec:eval} we perform an experimental evaluation. Finally, in section \ref{sec:disc} we conclude our work and discuss future research directions.
%
%
\begin{figure}[ht!]
\centering
  \includegraphics[width=0.8\linewidth]{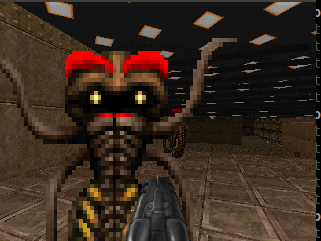}
\caption{Screen-shot from a Doom environment. The agent has to kill the Monster in order to survive.}
\label{fig:doom}
\end{figure}
\section{Theoretical background}
\label{sec:theor}
\subsection{Reinforcement Learning and the Actor-Critic Algorithm} 
A decision making problem can be defined as a Markov Decision Process (MDP), which is described from the tuple $(S, A, R, T)$, where $S$ is the set of states, $A$ is the set of actions, $R:S\times A \times S \rightarrow \mathbb{R}$ is the reward function and $T(s'|s,a) : S \times A \times S \rightarrow [0,1]$ is the transition function. We define the policy function, which is the function that decides the action based on the state $\pi : S \rightarrow A$. In every MDP, given a policy $\pi$, we define the state value function as the expected sum of discounted rewards from the current state until the end of the episode $V^{\pi}(s) = \mathbb{E}_{\pi}[\sum_{t=0}^{T} \gamma^{t}r_t | s_t = s]$, where $\gamma$ is called discount factor and takes values between $0$ and $1$. The goal of reinforcement learning is to compute the policy function that maximizes the expected sum of discounted rewards $\mathbb{E}_{\pi}[\sum_{t=0}^{T} \gamma^{t}r_t]$. In our work, we use the actor-critic Konda \textit{et al.} \cite{b3} algorithm in order to compute the policy of our agent.
Using the policy gradient theorem and the definition of the state value function $V$, we can compute the parameters of the policy function by minimizing the error below.
%
%
 $$L_{\pi} = - (r + \gamma V(s') - V(s))log(\pi(a|s))$$
 For the approximation of the state value function we will minimize the standard TD error: $$L_{V} = \frac{1}{2}(r + \gamma V(s') - V(s))^2$$
For the approximation of the policy and the state value function we will use the deep architecture that was proposed by Mnih \textit{et al.} \cite{b2}. The policy and the state value networks share all the layers except the output layer; the policy output layer is a softmax function and the state value output layer is liner with one node.
\subsection{Asynchronous framework}
The asynchronous framework was proposed by Mnih \textit{et al.} \cite{b2} in order to combine on-policy algorithms with deep learning. There is a global neural network which is responsible for learning the policy and the state value function. Furthermore, it creates a number of parallel processes and each process has a copy of the global network and interacts with the environment independently from the other processes. After a given number of (local) steps each process updates asynchronously the weights of the global network. The combination of actor-critic and the asynchronous framework resulted in the asynchronous advantage actor-critic (a3c) (Mnih \textit{et al.} \cite{b2}). The a3c algorithms led not only to better results in a plethora of environments but also significantly reduced the training steps.
\subsection{Unsupervised auxiliary tasks}

The unsupervised auxiliary tasks were used in combination with a3c algorithm (Jaderberg \textit{et al.} \cite{b4}) in order to enhance the learning ability of the neural network that was used for the policy approximation. In order to use the auxiliary tasks, an experience replay buffer is included. In this work we used three auxiliary tasks, which are described below:
\begin{itemize}
\item \textbf{Value function replay:} We uniformly sample from the experience replay and we update again the state value function using the temporal difference error:$L_{VR} = \frac{1}{2}(r + \gamma V(s') - V(s))^2$.
\item \textbf{Reward prediction:} In the deep network, which is responsible for the policy and the value function approximation, we add one more \textit{softmax} output. This output predicts the sign of the reward based on the last three frames. As a result there are three output nodes; one for zero reward, one for positive reward and one for negative reward. In order to avoid the class imbalance problem, we sample from the experience replay in a way that ensures that half of the samples have zero reward and the rest of them have negative or positive reward.
\item \textbf{Pixel change:} In the deep network, a deconvolutional layer is added. This layer is responsible for encouraging the agent to maximize the difference between two consecutive frames, due to the fact that changes in the sensory input many times lead to rewarding events. For this purpose, we consider a pseudo-reward - the absolute difference between two consecutive frames - and we maximize it using the Peng  {et al.} \cite{b6} n-step Q-learning algorithm.
%
%
%
%

\end{itemize}
\section{Relevant work}
Recently, there has been a great number of works trying to address the problem of visual Doom. Wu \textit{et al.} \cite{b5} used a3c and curriculum learning  in order to create a Doom agent. Lample \textit{et al.} \cite{b7} used a divide and conquer method in order to divide the action space. Then, they trained one agent for firing at the enemy monsters, the action agent, and one agent for navigating in the mazes of the game, the navigation agent. In order to decide, which agent had to act at each timestep they tried to detect if there is an enemy monster in the current frame. If an enemy exists, the action agent will act, otherwise the navigation agent will act.  In the next section, we will discuss our method, which is based on Lample \textit{et al.} \cite{b7} method, using a different way in order to determine which agent is going to act at each timestep.
%
%

\label{sec:relev_work}

\section{Methodology}
\label{sec:our_work}

In our work, we used a similar approach as Lample \textit{et al.} \cite{b7}. We separated the action space and we trained an agent for firing at the enemy monsters and one for navigating in the complex mazes and gathering objects. Additionally, from the set of the available moves we only kept five (5) for the action agent (FIRE, MOVE\_FORWARD, TURN\_RIGHT, TURN\_LEFT, MOVE\_BACKWARD) and three (3) for the navigation agent (MOVE\_FORWARD, TURN\_RIGHT, TURN\_LEFT). This is the minimum set of necessary actions for our agent. By reducing the number of actions, we significantly reduce the complexity of our system and the algorithms converge faster to a good policy. Both agents are trained using the UNREAL (Jaderberg \textit{et al.} \cite{b4}) algorithm, which is the combination of the a3c and the auxiliary tasks.

We used the same architecture that was proposed by Jaderberg \textit{et al.} \cite{b4}. The input is a RGB image with dimensions $84\times 84 \times 3$. The network has two convolutional layers with 16 $8 \times 8$ filters and 32 $4 \times 4 $ filters respectively. After the convolutional layers, a fully connected layer with 256 nodes is connected. All three layers use ReLU for activation function. Finally, the last hidden layer is a LSTM with 256 nodes and it is unrolled for 20 steps. The policy output is a \textit{softmax} function, while the state value output is  a linear function. At the output of the second convolutional layer we add a fully connected layer with 3 softmax output nodes. This layer is responsible for the reward prediction auxiliary task. For the pixel change task we connect a ReLU fully connected layer  and a deconvolutional layer in the output of the LSTM.

The agents try to minimize the cumulative error of the individual components; the policy error $L_\pi$, the value function error $L_{V}$, the value function replay error $L_{VR}$, the reward prediction error $L_{RP}$ and the pixel change error $L_{PC}$. In order to minimize the error below we used a shared \textit{RMSProp} optimizer (Mnih \textit{et al.} \cite{b2}).
$$ L = L_\pi +  L_{V} + \lambda_{VR} L_{VR} + \lambda_{RP} L_{RP} + \lambda_{PC} L_{PC}$$

We used the \textit{Vizdoom} (Kempka \textit{et al.} \cite{b8}) platform for training and evaluating our agents. This platform provides a large number of environments and game settings for training reinforcement learning agents based on visual input. The platform allows to shape the reward of the environment. This is necessary in order to efficiently train our agents. Without the reward shaping it would be impossible to train our agents due to the sparse reward of the environment. The reward shaping for the action and the navigation agent can be seen in the table below:

\begin{table}[h!]
  \centering
  \caption{The shaped rewards for the action and the navigation agent}
  \begin{tabular}{|c|c|c|c|c|c|}
    	\hline
       Agent & kill & death & missed shoot & lost health & object gathered \\
        \hline
Action  & +1 & -1 & -0.02  & -0.06  & +0.3   \\ 
\hline
Navigation  &  & -1 &   & -0.1  & +0.5   \\ 

		\hline
        
  \end{tabular}

  \label{table:reward_shape}
\end{table}

Having defined our action and our navigation agent, we now have to decide which agent is going to act given a frame. For this purpose, we will use one of the auxiliary tasks of the navigation agent; the reward prediction. From the table \ref{table:reward_shape}, we can understand that the navigation agent receives a negative reward only when it loses health or it is killed. From this we can conclude that there is an enemy close to our agent when it receives a negative reward. As a result, when there is a prediction for negative reward from the navigation agents, there is an enemy in the given frame. Otherwise, there is  a prediction for zero or positive reward. Experimentally, the combined agent performs better if the action agent acts when there is a prediction for positive reward. Therefore, if there is a prediction for positive or negative reward, the action agent will act, otherwise the navigation agent will act. 

\section{Evaluation}
\label{sec:eval}
Below you can find the experimental setup and the results both for training and testing.

\subsection{Experimental setup for training the agents}

We train both action and navigation agents using eight (8) parallel threads on four (4) different Doom maps. As a result two (2) actor learner threads are trained for each one of the four (4) maps. In order to increase the learning capacity of our agents, we trained both of them in environments with dense rewards for 80 million global steps. Thus, the action agent was trained in environments that have a large number of enemy monsters, while the navigation agent was trained in environments that have a plethora of objects to pick up.

\begin{figure}[ht!]
\centering
  \includegraphics[width=\linewidth]{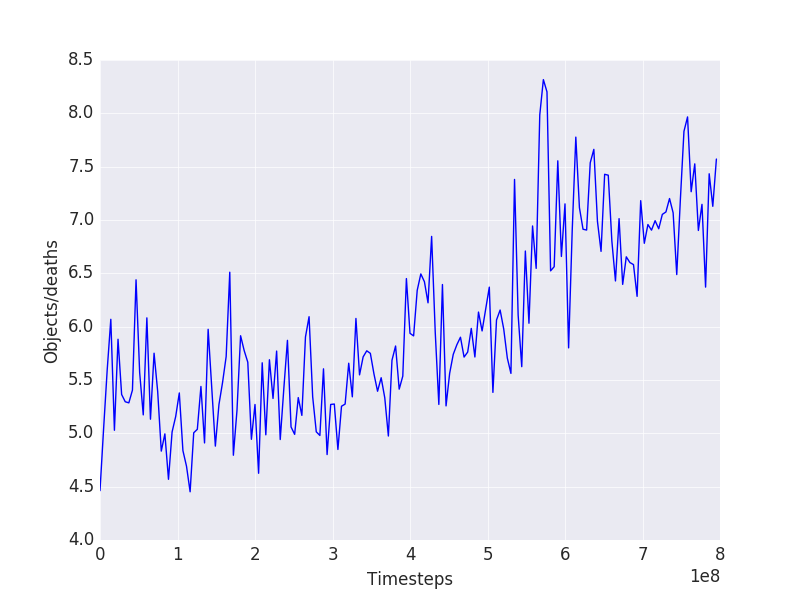}
\caption{Object-death ratio for the navigation agent during training.}
\label{fig:nav_training}
\end{figure}

\begin{figure}[ht!]
\centering
  \includegraphics[width=\linewidth]{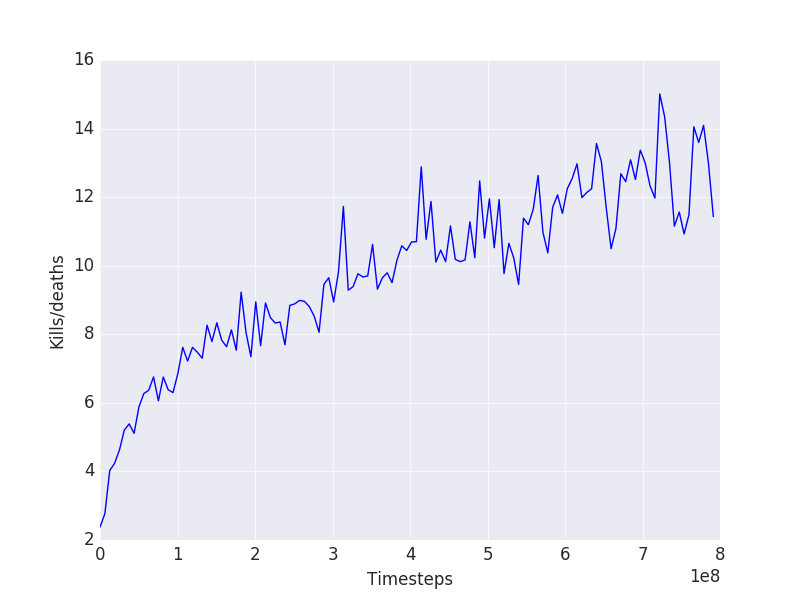}
\caption{Kill-death ration for the action agent during training.}
\label{fig:act_training}
\end{figure}

We need to formally define a set of measurements in order to evaluate our work. Since the reward that the agent receives from the game is based on a variety of the game's feature, we considered the following measurements: The goal of the action agent is to maximize the number of kills, while minimizing the number of deaths. Similarly, the goal of navigation agent is to maximize the number of objects it picks up, while minimizing the number of deaths. In the Figures \ref{fig:nav_training} and \ref{fig:act_training}, one can find the the object-death ratio and the kill-death ratio for the navigation and the action agent respectively.

\subsection{Experimental setup for testing the agents}
%
%

After evaluating our agents during their training, we tested the action agent and the combined agent both in known and unknown environments. We evaluated both agents in the four (4) maps that were used for training and two (2) unknown map. In the known maps, we severely decreased the number of enemy monsters compared to those that were used for training. The testing episode duration is 15 minutes in the known maps and in the second unknown map. In the first unknown map the episode terminates when the agent is killed. You can find the results in the table below:

\begin{table}[h]
	\centering
    	\caption{Results of the action agent and the combined agent in a variety of environments}
	\begin{tabular}{|c|c|c|c|c|c|c|}
    	\hline
		 & \multicolumn{3}{c|}{Action agent} & \multicolumn{3}{c|}{Combined agent} \\
		\hline
		Map & kills & deaths & objects & kills & deaths & objects \\
		\hline
        Known 1 & \textbf{13.32} & \textbf{3.23} & 29.51 & 13.15 & 4.58 & \textbf{37.73}  \\
		\hline
        Known 2 & \textbf{13.68} & \textbf{3.32} & 28.23 & 12.44 & 4.77 & \textbf{67.21} \\
		\hline
        Known 3 & \textbf{14.7} & \textbf{3.07} & 62.4 & 7.57 & 6.53 & \textbf{116.8}  \\
		\hline
        Known 4 & \textbf{8.72} & \textbf{1.43} & \textbf{48.76} & 6.45 & 1.86 & 36.94 \\
		\hline
        Unknown 1 & 4.99 & 1 & 5.46 & \textbf{5.89} & 1 & \textbf{20.19} \\
		\hline
        Unknown 2 & 10.23 & 1.97 & 13 & \textbf{12.94} & \textbf{1.88} & \textbf{36.4} \\
		\hline
	\end{tabular}

    \label{table:testEval}
\end{table}

From Table \ref{table:testEval}, it can be observed that the combined agent performs better than the action agent in the unknown environments.  In the known environments the action agent tends to perform better than the combined agent in kills and deaths statistics. This is not surprising due to the fact that the action agent is trained in these environment in order to act independently. Finally, it is clear that the combined agent gathered far more objects than the action agent.  This happened because the primary goal of the action agent is to fire at the enemy monsters, while the combined agent aims both for firing at the enemy monsters and for gathering objects.
%
%

After presenting our results we evaluated their statistical significance due to the fact that Doom environments are stochastic with high variation. We performed t-tests between the values of the combined and the action agent for all the evaluation metrics in all six environments. Their p-values are depicted in the following table. Generally speaking, we can conclude that our results are statistical significant with 95\% confidence (p-value below 0.05).

\begin{table}[h]
	\centering
    	\caption{The p-values of the t-test. Asterisk denotes statistical significance with 95\% confidence.}
	\begin{tabular}{|c|c|c|c|}
  
		\hline
		Map & kills & deaths & objects \\
		\hline
        Known 1 & 0.845 & 0.038*  & 0.024* \\
		\hline
        Known 2 & 0.157 & 0.049* & $1.56 \times 10^{-11} $*\\
		\hline
        Known 3 & $8.17 \times 10^{-8}$* & $1.9 \times 10^{-4}$*  & $6.08 \times 10^{-8}$*\\
		\hline
        Known 4 & $6.38 \times 10^{-4}$* & 0.22 & 0.039* \\
		\hline
        Unknown 1 & 0.037* & N/A & $1.9 \times 10^{-29}$ *\\
		\hline
        Unknown 2 & $8.17 \times 10^{-3}$* & 0.77 & $5.7 \times 10^{-18}$*  \\
		\hline
	\end{tabular}

    \label{table:testEval_pvalues}
\end{table}

\section{Discussion and future work}
\label{sec:disc}

In this work, we proposed a system (agent) for learning to efficiently play the game of Doom. Our agent, uses a divide and conquer method based on Lample \textit{et al.} \cite{b7} and a deep architecture for reinforcement learning based on Jaderberg \textit{et al.} \cite{b4}. Lample \textit{et al.} \cite{b7} method requires high-level informations from the environment in order to be used, like the existence or not of an enemy or an object in a frame. Our method requires only the reward signal.

In the future, we would like to evaluate how the divide and conquer method could be applied in other similar tasks. Additionally, we consider that the unsupervised auxiliary tasks are an interesting method, both for enhancing the learning capabilities of the agent and for learning features of the trained system. As a result, we are confident that the divide and conquer method combined with the auxiliary tasks could solve a  variety of problems.
%
%

\end{document}